\documentclass[10pt,twocolumn,letterpaper]{article}

\usepackage{iccv}
\usepackage{times}
\usepackage{epsfig}
\usepackage{graphicx}
\usepackage{amsmath}
\usepackage{amssymb}
\usepackage{multirow}
\usepackage{algorithm,epstopdf,algorithmic,subfigure,booktabs,ctable}
\usepackage{enumerate}

\usepackage[breaklinks=true,bookmarks=false]{hyperref}

\iccvfinalcopy 

\def\iccvPaperID{****} 

\begin{document}
	
	\title{Complementary Boundary Generator with Scale-Invariant Relation Modeling for Temporal Action Localization: Submission to ActivityNet Challenge 2020}
	
	\author{Haisheng Su, Hao Shao, Jinyuan Feng,  Zhenyu Jiang, Manyuan Zhang, \\
	    Wei Wu, Yu Liu, Hongsheng Li, Junjie Yan \\
		\\
		SenseTime Group Limited\\
		{\tt\small \{suhaisheng,shaohao,fengjinyuan,jiangzhenyu,zhangmanyuan,wuwei,liuyu,yanjunjie\}@sensetime.com}
		\\
	}
	
	\maketitle

	\begin{abstract}
		This technical report presents an overview of our solution used in the submission to ActivityNet Challenge 2020 Task 1 (\textbf{temporal action localization/detection}). Temporal action localization requires to not only precisely locate the temporal boundaries of action instances, but also accurately classify the untrimmed videos into specific categories. In this paper, we decouple the temporal action localization task into two stages (i.e. proposal generation and classification) and enrich the proposal diversity through exhaustively exploring the influences of multiple components from different but complementary perspectives. Specifically, in order to generate high-quality proposals, we consider several factors including the video feature encoder, the proposal generator, the proposal-proposal relations, the scale imbalance, and ensemble strategy. Finally, in order to obtain accurate detections, we need to further train an optimal video classifier to recognize the generated proposals. Our proposed scheme achieves the state-of-the-art performance on the temporal action localization task with \textbf{42.26} average mAP on the challenge testing set.
	\end{abstract}

	
	\section{Introduction}

\begin{figure}[t]
	\centering
	\setlength{\abovecaptionskip}{-0.1cm} 
	\includegraphics[width=1\columnwidth]{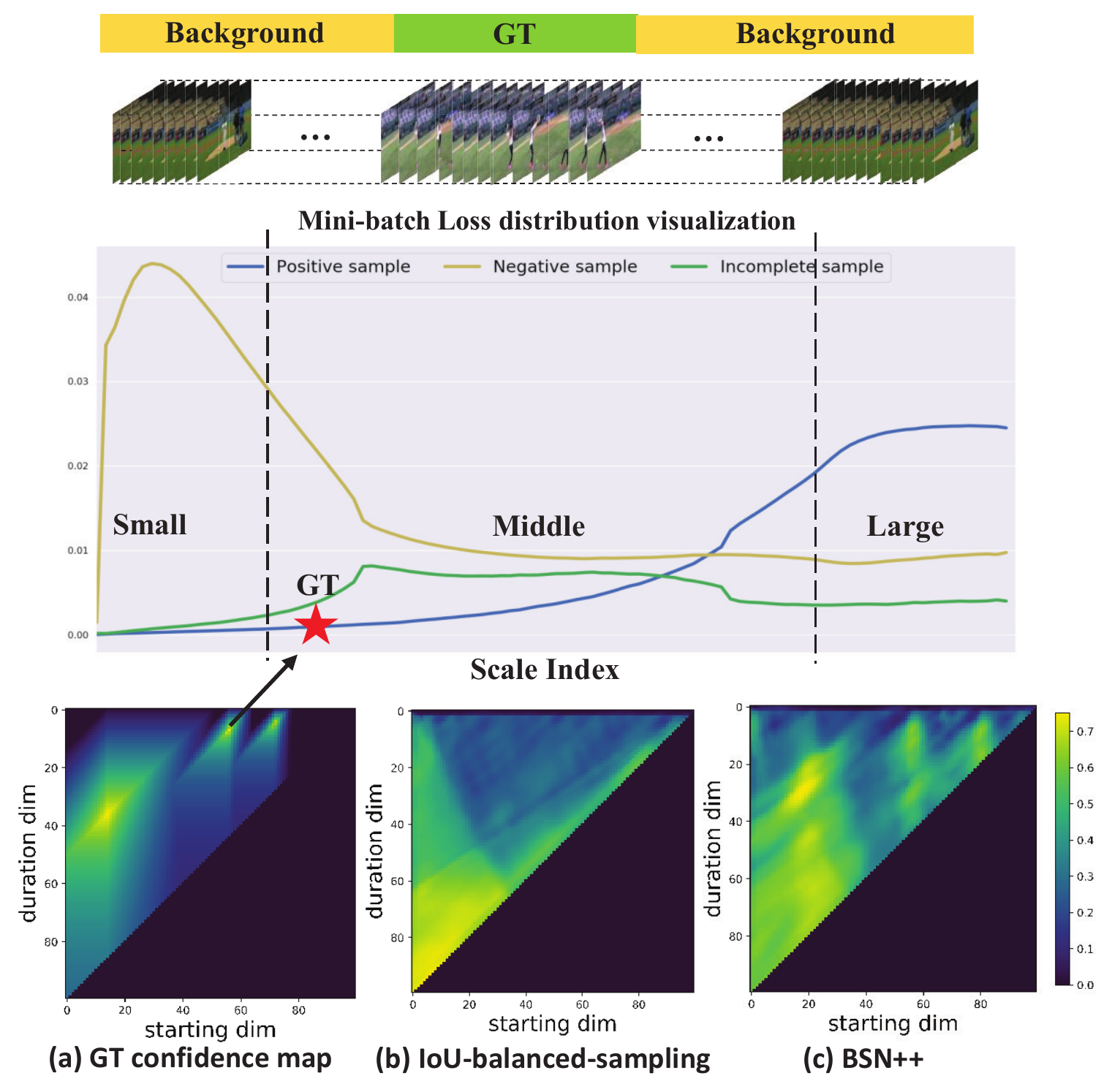} 
	\caption{(a) Given an untrimmed video containing several action instances of small scale, (b) IoU-balanced sampling is widely used to train the proposal confidence regressor, which still suffers from inferior quality owing to the imbalanced distribution of the temporal durations, resulting in the long-tailed proposal dataset. (c) BSN++ aims at generating high-quality proposal boundaries as well as reliable confidence scores with complementary boundary generator and scale-invariant proposal relation block. }
	\label{fig:overview}
	\vspace{-0.2cm}
\end{figure}

Temporal action detection task has received much attention from many researchers in recent years, which requires not only categorizing the real-world untrimmed videos but also locating the temporal boundaries of action instances. Akin to object proposals for object detection in images, temporal action proposal indicates the temporal intervals containing the actions and plays an important role in temporal action detection. It has been commonly recognized that high-quality proposals usually have two crucial properties: (1) the generated proposals should cover the action instances temporally with both high recall and temporal overlapping; (2) the quality of proposals should be evaluated comprehensively and accurately, thus providing a overall confidence for later retrieving step.

To cater for these two conditions and achieve high quality proposals, there are two main categories in the existing proposal generation methods \cite{sst_buch_cvpr17,gao2017turn,SSAD,SCNN}. The first type adopts the \textit{top-down} fashion, where proposals are generated based on sliding windows \cite{SCNN} or uniform-distributed anchors \cite{SSAD}, then a binary classifier is employed to evaluate confidence for the proposals. 
However, the proposals generated in this way are doomed to have imprecise boundaries though with regression. Under this circumstance, the other type of methods \cite{BSN,Y.Xiong,LinBMN} attract many researchers recently which tackle this problem in a \textit{bottom-up} fashion, where the input video is evaluated in a finer-level. \cite{BSN} is a typical method in this type which proposes the Boundary-Sensitive Network (BSN) to generate proposals with flexible durations and reliable confidence scores. Though BSN achieves convincing performance, it still suffers from three main drawbacks: (1) only the local details around the boundaries have been employed in BSN to predict boundaries, without taking advantage of the rich contexts through the whole video sequence; (2) the proposal-proposal relations fail to be considered for confidence evaluation; (3) the imbalance data distribution between positive/negative proposals and temporal durations is neglected.

To relieve these issues, we propose BSN++, for temporal action proposal generation. \textbf{(i)} To exploit the rich contexts for boundary prediction, we adopt the U-shaped architecture with nested skip connections. Meanwhile, the two optimized boundary classifiers share the same goals especially in detecting the sudden change from background to actions or learning the discriminativeness from actions to background, thus are complementary with each other. Under this circumstance, we propose the complementary boundary generator, where the starting classifier can also be used to predict the ending locations when the input videos are processed in a reversed direction during inference stage. In this way, we can achieve high precision without adding extra parameters. \textbf{(ii)} In order to predict the confidence scores of densely-distributed proposals, we design a proposal relation block aiming at leveraging both channel-wise and position-wise global dependencies for proposal-proposal relation modeling. \textbf{(iii)} To relieve the imbalance scale-distribution among the sampling positives as well as the negatives (see Fig. \ref{fig:overview}), we implement a two-stage re-sampling scheme consisting of the IoU-balanced (positive-negative) sampling and the scale-balanced sampling. The boundary map and the confidence map are generated simultaneously and jointly trained in a unified framework. In summary, the main contributions of our work are listed below in three-folds:

\begin{itemize}
	\item We revisit the boundary prediction problem and propose a complementary boundary generator to exploit both ``\textit{local and global}", ``\textit{past and future}" contexts for accurate temporal boundary prediction.
	
	\item We propose a proposal relation block for proposal confidence evaluation, where two self-attention modules are adopted to model the proposal relations from two complementary aspects. Besides, we devise a two-stage re-sampling scheme for equivalent balancing. 
	
	\item Thorough experiments are conducted to reveal the effectiveness of our method. Further combining with the existing action classifiers, our method can achieve the state-of-the-art temporal action detection performance.
	
\end{itemize}

\section{Related Work}
\subsection{Action Recognition}
Action recognition is an essential branch which has been extensively explored in recent years. Earlier methods such as improved Dense Trajectory (iDT) \cite{DT,iDT} mainly adopt the hand-crafted features including HOG, MBH and HOF. Current deep learning based methods \cite{C.Feichtenhofer,K.Simonyan,D.Tran,TSN} typically contain two main categories: the two-stream networks \cite{C.Feichtenhofer,K.Simonyan} capture the appearance features and motion information from RGB image and stacked optical flow  respectively; 3D networks \cite{D.Tran,p3d} exploit 3D convolutional layers to capture the spatial and temporal information directly from the raw videos. Action recognition networks are usually adopted to extract visual feature sequence from untrimmed videos for the temporal action proposals and detection task.

\subsection{Imbalanced Distribution Training} 
Imbalanced data distribution naturally exists in many large-scale datasets \cite{OpenImage,cityscapes}. Current literature can be mainly divided into three categories: (1) re-sampling, includes oversampling the minority classes \cite{Andrew} or downsampling the majority classes \cite{Gary}; (2) re-weighting, namely cost sensitive learning \cite{Kate,Cui}, which aims to dynamically adjust the weight of samples or different classes during training process. (3) In object detection task, the imbalanced data distribution is more serious between background and foreground for one-stage detector. Some methods such as Focal loss \cite{TsungYi} and online hard negative mining \cite{Abhinav} are designed for two-stage detector. In this paper, we implement the scale-balanced re-sampling upon the IoU-balanced sampling for proposal confidence evaluation, motivated by the mini-batch imbalanced loss distribution against proposal durations.

\subsection{Temporal Action Detection and Proposals}

Akin to object detection in images, temporal action detection also can be divided
into proposal and classification stages. Current methods train these two stages separately \cite{SCNN,G.Singh} or jointly \cite{sstad,SSAD}. As for proposal generation, top-down methods \cite{SCNN,SSAD} are mainly based on sliding windows or pre-defined anchors, while bottom-up methods \cite{Y.Xiong,BSN,LinBMN} first evaluate the actionness or boundary probabilities of each temporal location in a finer level. 
However, proposals generated in a local fashion of \cite{Y.Xiong} cannot be further retrieved without confidence scores evaluated from a global view. And probabilities sequence generated in \cite{BSN,LinBMN,LiuMulti} is sensitive to noises, causing many false alarms. Besides, proposal-proposal relations fail to be considered for confidence evaluation. Meanwhile, the imbalanced distribution among the proposals remains to be settled. To address these issues, we propose BSN++, which is unique to previous works in three main aspects: (1) we revisit the boundary prediction task and propose a complementary boundary predictor to exploit rich contexts together with bi-directional matching strategy for boundary prediction; (2) we propose a proposal relation block for proposal-proposal relations modeling; (3) two-stage re-sampling scheme is designed for equivalent balancing.

\begin{figure}[t]
	\centering
	\setlength{\abovecaptionskip}{-0.1cm} 
	\includegraphics[width=0.4\textwidth]{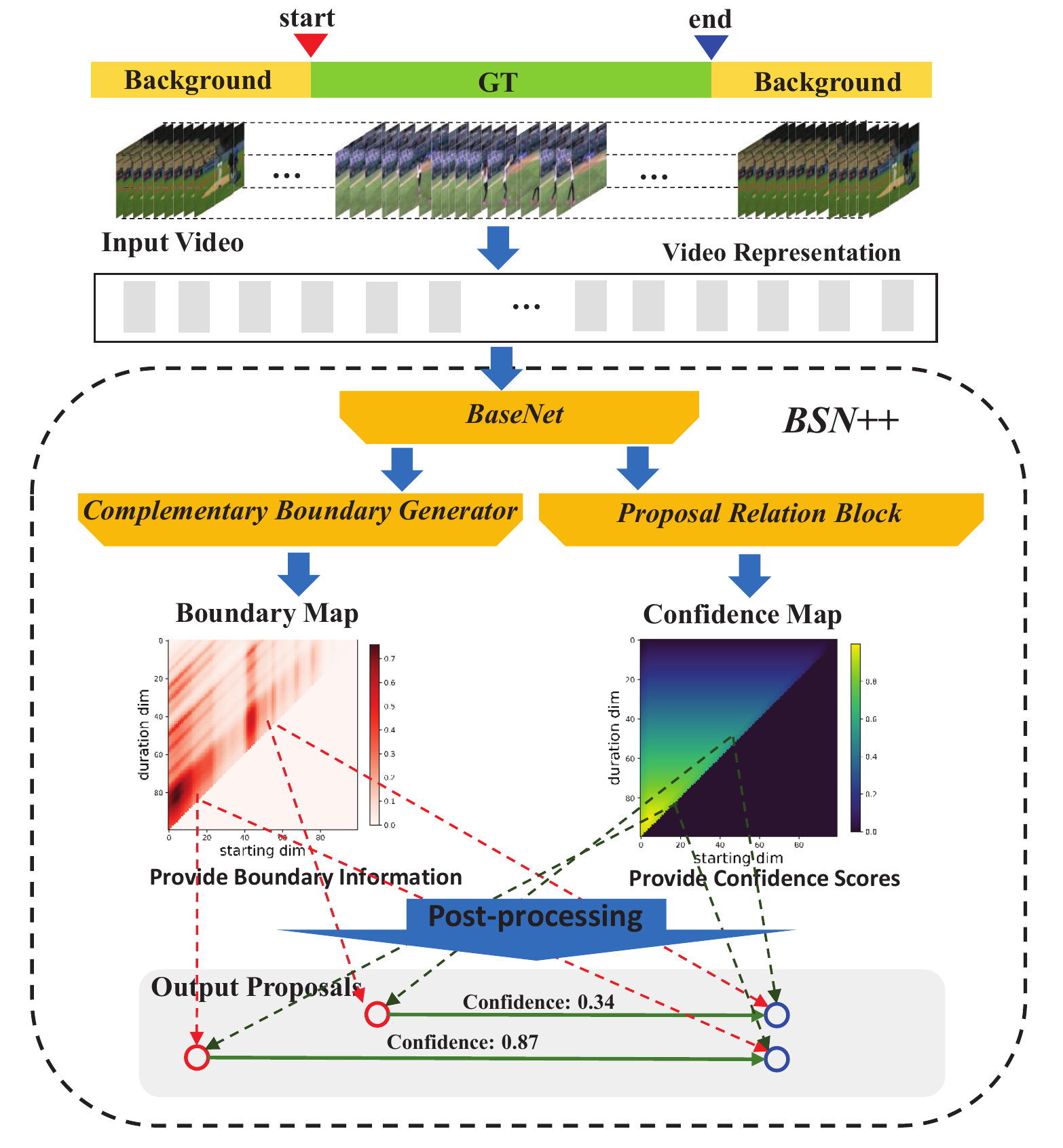} 
	\caption{The framework of BSN++. Given an untrimmed video, two-stream network is adopted to extract visual features. Then BSN++ can densely evaluate all proposals by producing the boundary map with a complementary boundary generator and the confidence map with a proposal relation block simultaneously.}
	\label{fig:framework}
	\vspace{-0.4cm}
\end{figure}

\section{Our Approach}


\subsection{Problem Definition}

Denote an untrimmed video sequence as $ \mathbf{U} = \{\mathbf{u}_{t}\}_{t=1}^{l_{v}}, $ where $ \mathbf{u}_{t} $ indicates the $ t $-th frame in the video of length $ l_{v} $. A set of action instances $ \mathbf{\Psi}_{g} = \{\mathbf{\varphi}_{n} = (t_{n}^{s}, t_{n}^{e})\}_{n=1}^{N_{g}} $ are temporally annotated in the video $  \mathbf{S}_{v} $, where $ N_{g} $ is the number of ground truth action instances, and $ t_{n}^{s}, t_{n}^{e} $ are the starting time and ending time of the action instance $ \varphi_{n} $ respectively. During training phase, the $ \mathbf{\Psi}_{g} $ is provided. While in the testing phase, the predicted proposal set $ \mathbf{\Psi}_{p} $ should cover the $ \mathbf{\Psi}_{g} $ with high recall and high temporal overlapping.

\subsection{Video Feature Encoding}

Before applying our algorithm, we adopt the two-stream network \cite{K.Simonyan} in advance to encode the visual features from raw video as many previous works \cite{BSN,Gao2018CTAP,Su2018Cascaded}. This kind of architecture has been widely used in many video analysis tasks\cite{SSAD,SSN,CBR}. Concretely, given an untrimmed video $ \mathbf{S}_{v} $ which contains $ l_{v} $ frames, we process the input video in a regular interval $ \sigma $ for reducing the computational cost. We concatenate the output of the last FC-layer in the two-stream network to form the feature sequence $ \mathbf{F}=\{\mathbf{f}_{i}\}_{i=1}^{l_{s}}$, where $ l_{s} = l_{v}/\sigma $. Final, the feature sequence $ \mathbf{F}$ is used as the input of our BSN++.

\subsection{Proposed Network Architecture: BSN++}

In contrast to the previous BSN \cite{BSN}, which consists of multiple stages, BSN++ is designed to generate the proposal map directly in an end-to-end network. To obtain the proposal map, BSN++ first generates the boundary map which represents the boundary information and confidence map which represents the confidence scores of densely distributed proposals. As shown in Fig. \ref{fig:framework}, BSN++ model mainly contains three main modules:  \textit{Base Module} handles the input video features to perform temporal information modeling, then the output features are shared by the two following modules. \textit{Complementary Boundary Generator} processes the input video features to evaluate the starting and ending probabilities sequence, using a nested U-shaped encoder-decoder network; \textit{Proposal Relation Block} aims to model the proposal-proposal relations with two self-attention modules responsible for two different but complementary dependencies.

\noindent
\textbf{Base Module.} The goal of this module is to handle the extracted video representations for temporal information modeling, which serves as the backbone of the following two branches. It mainly includes two 1D convolutional layers, with 256 filters, kernel size 3 and stride 1, followed by a ReLU activation layer. Since the length of videos is uncertain, we truncate the video sequence into a series of sliding windows. The detailed of data construction is illustrated in Section 4.1.

\begin{figure}[t]
	\centering
	\setlength{\abovecaptionskip}{-0.1cm} 
	\includegraphics[width=0.5\textwidth]{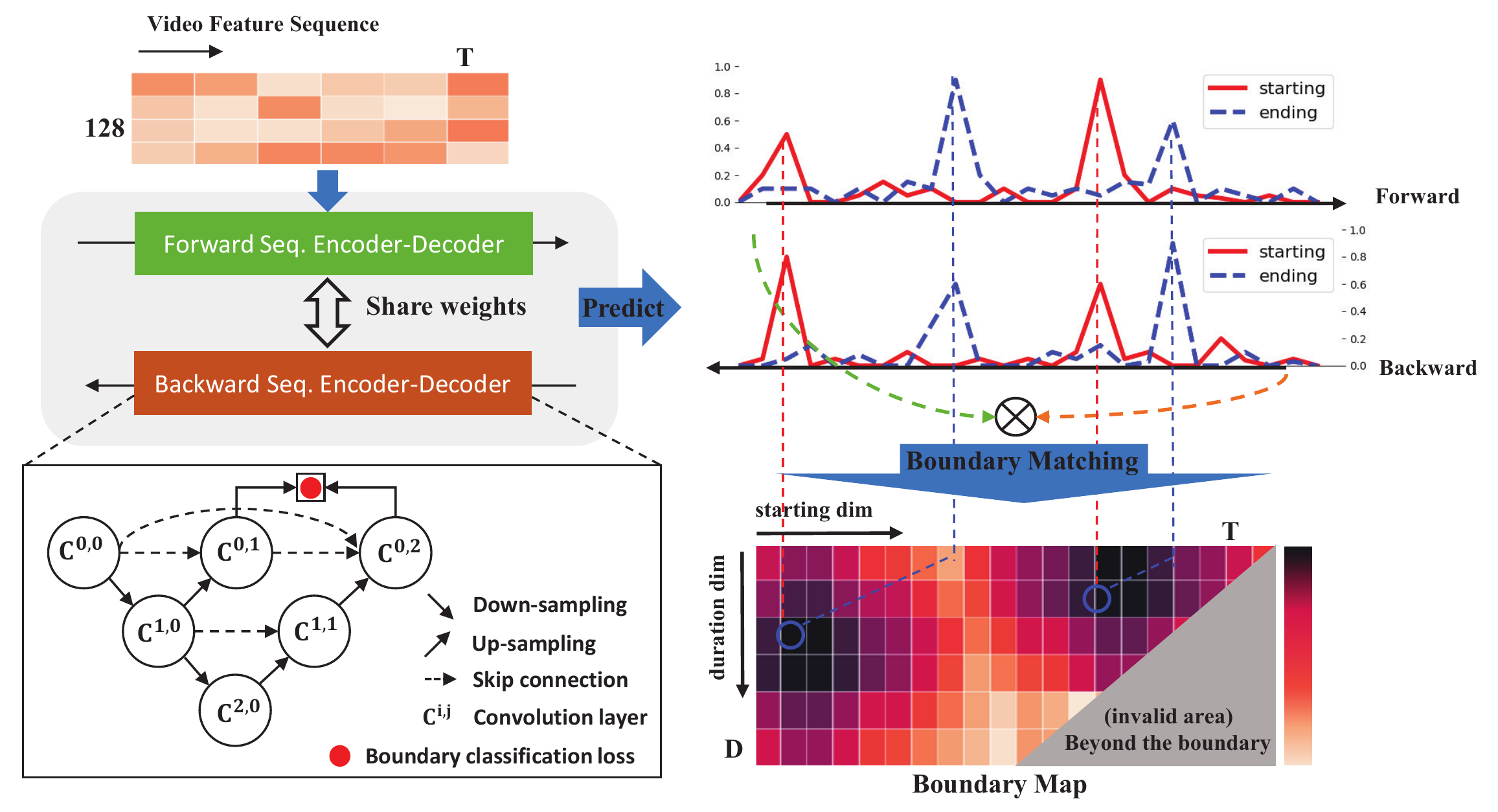} 
	\caption{Illustration of the complementary boundary generator. U-shaped encoder-decoder networks with dense skip connections are adopted to capture both \textit{local and global} contexts for accurate boundary prediction. During inference stage, our starting/ending classifiers are utilized to predict the ending/starting locations in a backward order, thus \textit{past and future} contexts can be further exploited without adding extra parameters. The two networks share the same weights. Then we fuse the two-passes boundary probabilities sequence and further construct the boundary map through matching each starting and ending locations into pairs. }
	\label{fig:TEM}
	\vspace{-0.4cm}
\end{figure}

\noindent
\textbf{Complementary Boundary Generator.} This module takes the output features of base module as input and perform boundary prediction. Inspired by the success of U-Net \cite{unet,Unet++} used in image segmentation, we design our boundary generator as Encoder-Decoder networks because this kind of architecture is able to capture both high-level \textit{global} context and low-level \textit{local} details at the same time. As shown in Fig. \ref{fig:TEM}, each circle represents a 1D convolutional layer with 512 filters and kernel size 3, stride 1, together with a batch normalization layer and a ReLU layer except the prediction layer. To reduce over-fitting, we just add two down-sampling layers to expand the receptive fields and the same number of up-sampling layers are followed to recover the original temporal resolutions. Besides, deep supervision (shown {\color{red}red}) is also performed for fast convergent speed and nested skip connections are employed for bridging the semantic gap between feature maps of the encoder and decoder prior to fusion.


We observe that the starting classifier learns to detect the sudden change from background to actions and vice versa. Hence, the starting classifier can be regarded as an another ``ending classifier'' when processes the input video in a reversed direction, thus the bi-directional prediction results are complementary. With this observation, we implement the bidirectional boundary matching step, which involves a forward pass and a backward pass. As for the forward pass, we use the aforementioned encoder-decoder network to predict the starting heatmap $ \overrightarrow{\mathbf{H}}^{s}=\{\overrightarrow{\mathbf{h}}_{i}^{s}\}_{i=1}^{l_{s}} $ and ending heatmap $ \overrightarrow{\mathbf{H}}^{e}=\{\overrightarrow{\mathbf{h}}_{i}^{e}\}_{i=1}^{l_{s}} $ respectively, where $ \mathbf{h}_{i}^{s}$ and $\mathbf{h}_{i}^{e}$ indicate the starting and ending probabilities of the $ i $-th snippet respectively. As for the backward pass, we feed the input feature sequence in a reversed order to the identical Encoder-Decoder network. It is expected to predict the high scores at the corresponding boundary locations as well. Similarly, we can obtain the starting heatmap $ \overleftarrow{\mathbf{H}}^{s}$ and ending heatmap $ \overleftarrow{\mathbf{H}}^{e}$ in the backward pass. 

After the two-passes, in order to select the boundaries of high scores, we fuse the two pairs of heatmaps to yield the final heatmaps:
\begin{equation}
	\mathbf{H}^{s} = \{\sqrt{\overrightarrow{\mathbf{h}_{i}^{s}} \times \overleftarrow{\mathbf{h}}_{i}^{s}}\}_{i=1}^{l_{s}},
	\mathbf{H}^{e} = \{\sqrt{\overrightarrow{\mathbf{h}}_{i}^{e} \times \overleftarrow{\mathbf{h}}_{i}^{e}}\}_{i=1}^{l_{s}}
\end{equation}

With these two boundary points heatmaps, we can further construct the boundary map $\mathbf{M}^{b} \in R^{1\times D\times T} $ which can represent the boundary information of all densely distributed proposals, where $ T $ and $ D $ are the length of the feature sequence and maximum duration of proposals separately:
\begin{equation}
	\mathbf{M}_{j,i}^{b} = \{\{\mathbf{h}_{i}^{s} \times\mathbf{h}_{i+j}^{e}\}_{i=1}^{T}\}_{j=1}^{D},   i + j < T
\end{equation}

\begin{figure}[t]
	\setlength{\abovecaptionskip}{-0.1cm} 
	\begin{center} 
		\includegraphics[width=1\linewidth]{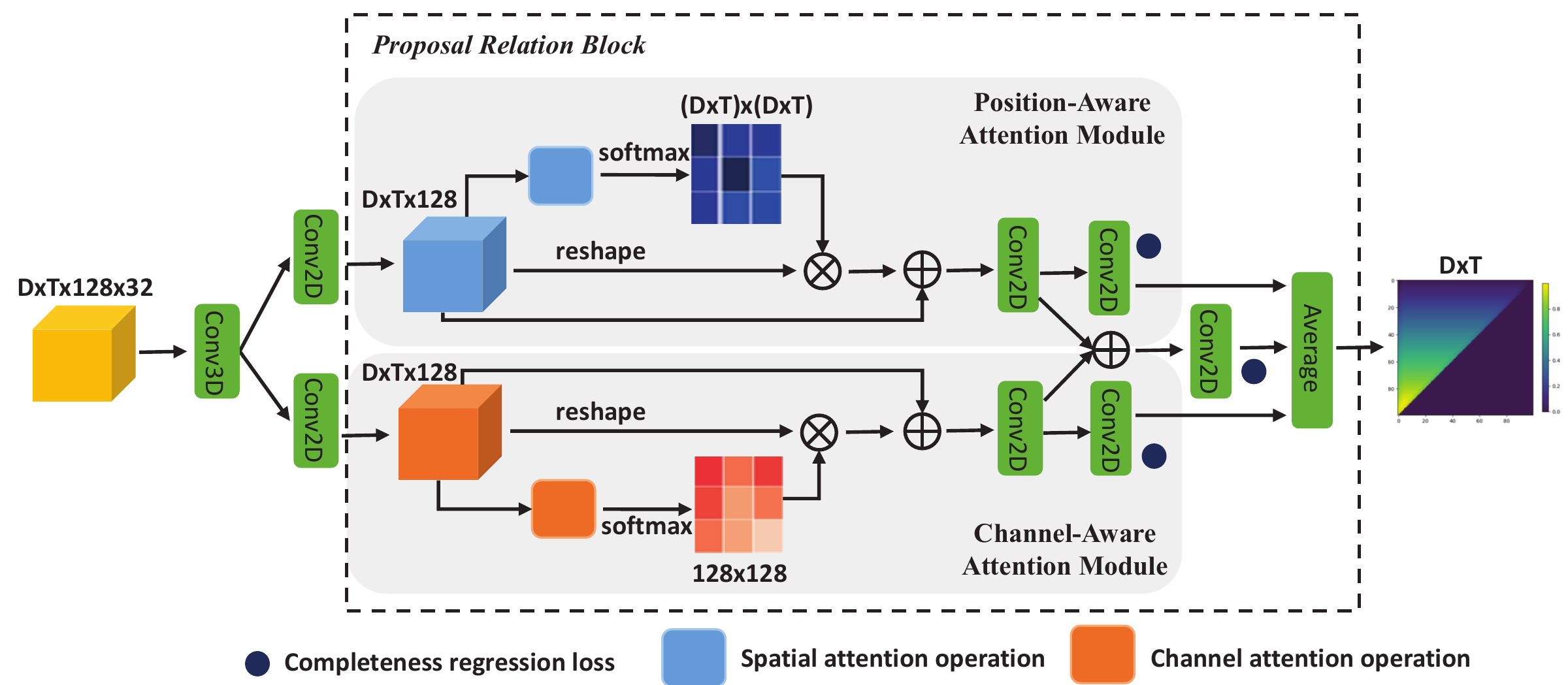}
	\end{center}
	\caption{Illustration of the proposal relation block. After generating the proposal feature maps, two complementary branches are followed to model the proposal relation separately. In the upper branch, the position-aware attention module aims to leverage \textit{global} dependencies. While in the lower branch, the channel-aware attention module aims to attend to the discriminative features by channel matrix calculation. Finally, we aggregate the outputs from the three branches for \textit{pixel-level} confidence prediction.}
	\label{fig:PEM}
	\vspace{-0.3cm}
\end{figure}   

\noindent
\textbf{Proposal Relation Block.} The goal of this block is to evaluate the confidence scores of dense  proposals. Before performing proposal-proposal relations, we follow the previous work BMN \cite{LinBMN} to generate the proposal feature maps as $\mathbf{F}^{p} \in R^{D\times T\times 128\times N } $. N is set to 32. Then the proposal feature maps are fed to a 3D convolutional layer with kernel size $ 1\times 1\times 32 $ and 512 filters, followed by a ReLU activation layer. Thus the reduced proposal features maps are $\widehat{\mathbf{F}^{p}} \in R^{D\times T\times 512} $. The proposal relation block consists of two self-attention modules as follows.

\textbf{Position-aware attention module.} As illustrated in Fig. \ref{fig:PEM}, given the proposal feature maps $\widehat{\mathbf{F}^{p}}$, we adopt the similar self-attention architecture as Non-local network \cite{nnn}, where the proposal feature maps are fed into a convolutional layer separately to generate 
two new feature maps $ A $ and $ B $ for spatial matrix multiplication with reshape and transpose operations. And then a Softmax layer is applied to calculate the position-aware attention $ P^{A}\in R^{L\times L} $, where $ L=D\times T $:
\begin{equation}
P^{A}_{j,i} = \dfrac{exp(A_{i}\cdot B_{j})}{\sum_{i=1}^{L}exp(A_{i}\cdot B_{j})} ,
\end{equation}
where $ P^{A}_{j,i} $ indicates the attention of the $ i^{th} $ position on the $ j^{th} $ position. The more similar of the features means the closer relation between the two proposals. Finally, the attended features are further weighted summed with the proposal features and fed to the convolutional layers for confidence prediction.

\textbf{Channel-aware attention module.} In contrast to the position-aware attention module, this module directly performs channel-wise matrix multiplication in order to exploit the inter-dependencies among different channels, which can help enhance the proposal feature representations for confidence prediction. The process of attention calculation is the same as the former module except for the attended dimension. Similarly, the attended features after weighted summed with the proposal features are further captured by a 2D convolutional layer to generate the confidence map $ \mathbf{M}^{c} \in R^{D\times T} $. We also aggregate the outputs of the two attention modules for proposal confidence prediction, and finally we fuse the predicted confidence maps from the three branches for a better performance.


\subsection{Re-sampling}

Imbalanced data distribution can affect the model training especially in the long-tailed dataset. In this paper, we revisit the positive/negative samples distribution for improving the quality of proposal confidence prediction and design a proposal-level re-sampling method to improve the performance of training on the long-tailed dataset. Our re-sampling scheme consists of two stages aiming at not only balancing the positives and negatives proposals, but also balancing the temporal duration of the proposals.
\vspace{-0.4cm}
\subsubsection{IoU-balanced sampling} As shown in Fig. \ref{fig:overview}, we can see from the mini-batch loss distribution that the number of positives and negatives differs greatly which dooms to bias the training model without effective measures. Previous works usually design a positive-negative sampler (i.e. IoU-balanced sampler) to balance the data distribution for each mini-batch, thus ensuring the ratio of positive and negative samples is nearly 1:1. However, we can also conclude from the Fig. \ref{fig:overview} that the scale of positives or negatives fails to conform the uniform distribution. Under this circumstance, we should consider how to balance the scales of proposals.
\vspace{-0.4cm}
\subsubsection{Scale-balanced re-sampling} To relieve the issue among long-tailed scales, we propose a second-stage positive/negative re-sampling method, which is upon the principle of IoU-balanced sampling. Specifically, define $ P_{i} $ as the number of positive proposals with the scale $ s_{i} $, then $ r_{i} $ is the positive ratio of $ s_{i} $:
\begin{equation}
\begin{split}
   & r_{i} = \dfrac{P_{i}}{\sum_{j=1}^{N_{s}} P_{j}} , \\
   & r^{'}_{i} = \left\{  
   \begin{aligned}
             & \lambda \ast exp^{(\dfrac{r_{i}}{\lambda} - 1)} \quad & (0 < r_{i} \leq \lambda ) ,  \\  
             &  r_{i} \quad &(\lambda < r_{i} \leq 1 ) , 
   \end{aligned}  
\right. 
\end{split}
\end{equation}
where $ N_{s} $ is the number of pre-defined normalized scale regions (i.e. $[0 - 0.3, 0.3 - 0.7, 0.7 - 1.0]$). Then we design a positive ratio sampling function, the resulting ratio $  r^{'}_{i}$ is bigger than $  r_{i}$ for proposal scale with a frequency lower than $ \lambda$, where $ \lambda$ is a hyper-parameter which we set to 0.15 empirically. Hence, we use the re-normalized $  r^{'}_{i}$ as the sampling probability of the specific proposal scale region $ s_{i} $ to construct the mini-batch data. As for the negative proposals, the same process is performed.

\section{Training and Inference of BSN++}

In this section, we will introduce the training and inference details of BSN++.

\subsection{Training}
\textbf{Overall Objective Function.} As described above, BSN++ consists of three main sub-modules. The multi-task objective function is defined as:
\begin{equation}
	L_{BSN++} = L_{CBG} + \beta\cdot L_{PRB} + \gamma\cdot L_{2}(\Theta),
\end{equation}
where $ L_{CBG} $ and $ L_{PRB} $ are the objective functions of the complementary boundary generator and the proposal relation block respectively, while $ L_{2}(\Theta)$ is a regularization term. $\beta $ and $ \gamma $ are set to 10 and 0.0001 separately to trade off the training process of two modules and reduce over-fitting. 

\noindent
\textbf{Training Data Construction.} Given the extracted feature sequence $\mathbf{ F }$ with length $ l_{s} $, we truncate $\mathbf{ F }$ into sliding windows of length $ l_{w} $ with 75\% temporal overlapping. Then we construct the training dataset as $ \Phi = \{\mathbf{ F }^{w}_{n}\}_{n=1}^{N_{w}} $, where $ N_{w} $ is the number of retained windows containing at least one ground-truth. 	

\noindent
\textbf{Label Assignment.} For the Complementary Boundary Generator (CBG), in order to predict the boundary probabilities sequence, we need to generate the corresponding label sequence $ \mathbf{G}_{s}^{w} $ and $ \mathbf{G}_{e}^{w} $ as in \cite{BSN}. Specifically, for each action instance $ \varphi_{g} $ in the annotation set $ \mathbf{\Psi}_{g}^{w} $, we denote it's starting and ending regions as $ [t_{g}^{s}-d_{\varphi}/10, t_{g}^{s}+d_{\varphi}/10] $ and $ [t_{g}^{e}-d_{\varphi}/10, t_{g}^{e}+d_{\varphi}/10] $ respectively, where $ d_{\varphi}= t_{g}^{e} - t_{g}^{s} $ is the duration of $ \varphi_{g} $. Then for each temporal location, if it lies in the starting or ending regions of any action instances, the corresponding label $ \mathbf{g}^{s} $ or $ \mathbf{g}^{e} $ will be set to 1. Hence the label sequence of starting and ending used in CBG are $ \mathbf{G}_{s}^{w}=\{\mathbf{g}_{i}^{s}\}_{i=1}^{l_{w}} $, $ \mathbf{G}_{e}^{w}=\{\mathbf{g}_{i}^{e}\}_{i=1}^{l_{w}} $ respectively.

For the Proposal Relation Block (PRB), we predict the confidence map $\mathbf{M}^{c} \in R^{D\times l_{w}} $ of all densely distributed proposals, where the point $ g^{c}_{j,i} $ in the label confidence map $\mathbf{M}_{g}^{c} = \{\{ g^{c}_{j,i} \}_{i=1}^{l_{w}}\}_{j=1}^{D} $ represents the maximum $ IoU $ (Intersection-over-Union) values  of proposal $ \varphi_{j,i} = [t_{s} = i, t_{e} = i+j] $ with all $ \varphi_{g} $ in $ \mathbf{\Psi}_{g}^{w} $.

\noindent
\textbf{Objective of CBG.} We follow \cite{BSN} to adopt the weighted binary logistic regression loss $ L_{bl} $ as the objective between the output of the evaluated boundary probabilities sequence and the corresponding label sequence. The objective is:
\begin{equation}
	L_{CBG} = \underbrace{\overrightarrow{L_{bl}^{s}} + \overrightarrow{L_{bl}^{e}}}_{forward},
\end{equation}
where $ \overrightarrow{L_{bl}^{s}} $ and $ \overrightarrow{L_{bl}^{e}} $ represent the $ L_{bl} $ between $ \overrightarrow{\mathbf{H}}^{s} $ and $ \mathbf{G}_{s}^{w} $, $ \overrightarrow{\mathbf{H}}^{e} $ and $ \mathbf{G}_{E}^{w} $ respectively in the forward pass. 

\noindent
\textbf{Objective of PRB.} Taking the constructed proposal feature maps $\mathbf{F}^{p} $ as input, our PRB will generate two types of confidence maps $\mathbf{M}^{cr}$ and $\mathbf{M}^{cc}$ for all densely distributed proposals as \cite{LinBMN}. The training objective is defined as the regression loss $ L_{reg} $ and the binary classification loss $ L_{cls} $ respectively:
\begin{equation}
	L_{PRB} =  L_{reg} + L_{cls} ,
\end{equation}
where the smooth-$ L_{1} $ loss \cite{Girshick2015Fast} is adopted as  $ L_{reg} $, and the points  $ g^{c}_{i,j} $ with value large than 0.7 or lower than 0.3 are regarded as positives and negatives respectively. And we ensure the scale and number ratio between positives and negatives to be near 1:1 by the two-stage sampling scheme described above.

\subsection{Inference}
During inference stage, our BSN++ can generate the boundary map $ \mathbf{M}^{b} $ based on the bidirectional boundary probabilities ($ \mathbf{H}^{s}$ and $\mathbf{H}^{e}$) and confidence map ($ \mathbf{M}^{cc} $ and $ \mathbf{M}^{cr} $). We form the proposal map $ \mathbf{M}^{p} $ directly by fusing the $ \mathbf{M}^{b} $ and $ \mathbf{M}^{c} $ with dot multiplication. Then we can filter the points with high scores in the proposal map $ \mathbf{M}^{p} $ as candidate proposals used for post-processing.

\noindent
\textbf{Score Fusion.} As described above, the final scores of proposals in $ \mathbf{M}^{p} $ involve the local boundary information and global confidence scores. Take the proposal $\varphi$ = $[t_{s}, t_{e}]$ for example, the combination of final score $ p_{\varphi} $ can be shown as:
\begin{equation}
	\begin{aligned}
		p_{\varphi} & = \mathbf{M}^{b}_{t_{e}-t_{s},t_{s}} \cdot \sqrt{\mathbf{M}^{cc}_{t_{e}-t_{s},t_{s}} \cdot \mathbf{M}^{cr}_{t_{e}-t_{s},t_{s}}}\\
	\end{aligned}
\end{equation}

\noindent
\textbf{Redundant Proposals Suppression.} With the above process, our BSN++ can generate the proposal candidates set as $ \mathbf{\Psi}_{p}= \{\varphi_{n}=(t_{s},t_{e}, p_{\varphi})\}_{n=1}^{N_{p}}$, where $ N_{p} $ is the number of proposals. Since the generated proposals may overlap with each other to various degrees, we conduct Soft-NMS \cite{Bodla2017} algorithm to suppress the confidence scores of redundant proposals. Final, the proposals set is $ \mathbf{\Psi}_{p}'= \{\varphi_{n}^{'}=(t_{s},t_{e}, p_{\varphi}^{'})\}_{n=1}^{N_{p}}$, where $ p_{\varphi}^{'} $ is the decayed score of proposal $ \varphi_{n}^{'} $. It should be noted that we also try Greedy-NMS in experiments for fair comparison.

\section{Experiments}

\subsection{Datasets and Setup}
\noindent
\textbf{Datasets.} \textbf{ActivityNet-1.3} \cite{Anet} is a large-scale video dataset for action recognition and temporal action detection tasks used in the ActivityNet Challenge from 2016 to 2020. It contains 19, 994 videos with 200 action classes temporally annotated, and the ratio of training, validation and testing sets is 1:1:2. 

\noindent
\textbf{Evaluation metrics.} For temporal action detection task, mean Average Precision (mAP) is a conventional evaluation metric, where Average Precision (AP) is calculated for each action category respectively. On ActivityNet-1.3, the mAP with \textit{tIoU} thresholds set $ \{0.5, 0.75, 0.95\} $ and the average mAP with \textit{tIoU} thresholds from 0.5 to 0.95 with a step size of 0.05 are all reported.

\noindent
\textbf{Implementation details.} For feature encoding, we adopt the two-stream network \cite{K.Simonyan} with the architecture described in \cite{TSN}, where ResNet network \cite{K.He} and BN-Inception network \cite{S.Ioffe} are used as the spatial and temporal networks respectively. During feature extraction, the interval $ \sigma $ is set to 16 on ActivityNet-1.3. Then we rescale the feature sequence of input videos to $ l_{w}=100 $ by linear interpolation following the previous \cite{BSN}, and the maximum duration $ D $ is also set to 100 to cover all action instances. We train our BSN++ from scratch using the Adam optimizer and the batch size is set to 16. And the initial learning rate is set to 0.001 for 7 epochs, then 0.0001 for another 3 epochs.

\setlength{\tabcolsep}{15pt}
\begin{table*}[t]
	\centering
	\caption{Action detection results on validation and testing set of ActivityNet-1.3 in terms of mAP@\textit{tIoU} and average mAP, where our proposals are combined with video-level classification results generated by \cite{xiong2016cuhk}.}
	\small
	\begin{tabular}{p{2.5cm}p{0.62cm}<{\centering}p{0.62cm}<{\centering}p{0.62cm}<{\centering}p{0.9cm}<{\centering}p{0.9cm}<{\centering}}
		\toprule
		\multicolumn{6}{c}{ {\bf ActivityNet-1.3}, mAP@$tIoU$}  \\
		\hline
		& \multicolumn{4}{c}{validation} & \multicolumn{1}{c}{testing} \\
		\hline
		Method  & 0.5  &  0.75  & 0.95  & Average  & Average \\
		\hline
		CDC\cite{CDC}    & 43.83  & 25.88  & 0.21   & 22.77  &  22.90 \\
		SSN\cite{SSN}    & 39.12 & 23.48  & 5.49  & 23.98 & 28.28 \\
		SSAD\cite{SSAD} & 44.39  & 29.65  & 7.09  & 29.17  & 32.26 \\
		BSN\cite{BSN} + \cite{xiong2016cuhk} & 46.45 & 29.96 & 8.02 & 30.03 & 32.87 \\
		BMN\cite{LinBMN} + \cite{xiong2016cuhk} & 50.07 & 34.78 & 8.29 & 33.85 & 36.42 \\
		\hline
		Ours - baseline & 51.27  & 35.70  & \textbf{8.33}  & 34.88  & 37.44 \\
		\textbf{+ improvement A} & - & - & -  & - & 40.21 \\
		\textbf{+ improvement B} & - & - & -  & - &  40.99 \\
		\textbf{+ improvement C} & - & - & -  & - & 41.97 \\
		\textbf{+ improvement D} & \textbf{64.25}  & \textbf{46.28}  & 8.18  & \textbf{44.16}  & \textbf{42.26} \\
		\bottomrule
	\end{tabular}
	\label{table_detection_anet}
\end{table*}

\subsection{Temporal Action Localization}
	The detection performance comparisons of our method with previous state-of-the-arts are shown in Table \ref{table_detection_anet}. We can observe that our BSN++ can achieve superior results than other approaches. During the period of challenge, we further conduct several improvements to promote the overall detection performance, which are listed as follow. Note that each improvement is performed upon the last one.
	
\begin{enumerate}[(A)]
	\item \textit{Video features}: In BSN++ baseline, we adopt the two-stream network \cite{K.Simonyan} pre-trained on ActivityNet-1.3 to encode the visual features of an input video. In order to maximize the quality and diversity of video features, we adopt many other ConvNet architectures pre-trained on Kinetics-700 dataset to further fine-tune on ActivityNet-1.3 with both modalities, including ResNet-50, BN-Inception, SlowFast-101~\cite{feichtenhofer2019slowfast}, pseudo-3D network~\cite{qiu2017learning}, TIN~\cite{shao2020temporal}, TSM~\cite{lin2019tsm} and ResNest-269~\cite{zhang2020resnest}, which are then adopted as visual encoders for feature extraction. Table \ref{feature} illustrates the exact list of two-stream encoders which we adopt to extract video representations. 
    We also use some of the above models to encode the features of HACS~\cite{zhao2019hacs} dataset, and adopt HACS's feature to pretrain our BSN++. To fuse these features for enriching proposal diversity, we train BSN++ with these encoders separately, and then fuse the output of boundary map and confidence map through weighted combination instead of averaging, where the normalized weights are calculated according to the relative performance of each feature.
	
  \setlength{\tabcolsep}{16pt}
  \begin{table*}[t]
  	\centering
  	\caption{List of two-stream encoders with different backbones used to extract video representations.}
  	\small
  	\begin{tabular}{p{1cm}p{4cm}<{\centering}p{4cm}<{\centering}}
  		\toprule
  		\textbf{ID} & \textbf{RGB Feature} & \textbf{Flow Feature}   \\
	 	\hline
		1 & TSM\_rgb\_ResNest101\_K600 & TSM\_flow\_ResNest50\_K600  \\
  		2 & TSM\_rgb\_ResNest269\_K600 & TSM\_flow\_ResNest269\_K600  \\
		3 & TSM\_rgb\_ResNest269\_ANet & TSM\_flow\_ResNest50\_ANet \\
		4 & TSN\_rgb\_BnInception\_K600 & TSN\_flow\_BnInception\_K600 \\
		5 & Slowfast101\_rgb\_K600 & Slowfast101\_flow\_K600 \\
		6 & Slowfast101\_rgb\_K700 & Slowfast101\_flow\_K700 \\
		7 & TIN\_rgb\_ResNet101\_K700 & TIN\_flow\_ResNet101\_K700 \\
		8 & P3D\_rgb\_ResNet152\_K600 & P3D\_flow\_ResNet152\_K400 \\
  		\bottomrule
  	\end{tabular}
  	\label{feature}
  \end{table*}

	\item \textit{Video classifier}: In this part, we have trained many models like SlowFast-101(8$\times$8), SlowFast-101(16$\times$8), TIN-101, TSM-101 which are pre-trained on Kinetics-700 dataset with both rgb and optical flow modalities. Finally, grid-search method is utilized to find the best weight for each model when ensemble the output logits from different models, including the video-level classification results generated by \cite{xiong2016cuhk}.
	
	\item \textit{Ensemble of multi-scale prediction results}: We rescale the video feature to different scales including 30, 80 and 100 during inference stage. Based on our observation on the validation set, for the videos less than 30 seconds, we adopt the output results with the input feature scaled to 30; for the videos longer than 30 seconds and less than 120 seconds, we take the output results with the input feature scaled to 80; while for the videos longer than 120 seconds, we take output detections with the input feature scaled to 100.
	
	\item \textit{Ensemble with BMN~\cite{LinBMN}}: In order to further promote the model diversity, we combine the detection results generated by BMN through concatenating directly. Note that BMN is performed on all video features we mentioned above to ensure the competitive performance. Finally, we conduct Soft-NMS algorithm to suppress redundant detections.
	
\end{enumerate}

	Finally, our ensembled method achieves \textbf{44.16} average mAP on the validation set of ActivityNet-1.3 and \textbf{42.26} average mAP on the testing server of ActivityNet Challenge.

	\section{Conclusion}
	In this challenge notebook, we have introduced our recent work, namely BSN++ for temporal action proposal generation. The complementary boundary generator takes the advantage of U-shaped architecture and bi-directional boundary matching mechanism to learn rich contexts for boundary prediction. To model the proposal-proposal relations for confidence evaluation, we devise the proposal relation block which employs two self-attention modules to perform global dependencies and inter-dependencies modeling. Meanwhile, we are the first to consider the imbalanced data distribution of proposal durations. Both the boundary map and confidence map can be generated simultaneously in a unified network. And we further introduce several improvements as well as ensemble strategies we conducted during the ActivityNet challenge 2020. All these improvements can further promote the detection performance obviously (+5\%), which also reveal the direction of how to make better temporal action proposal generation and localization.
	  

	{\small
		\bibliographystyle{ieee}
		\bibliography{egbib}
	}
	
\end{document}